*Article*

# Integral Images: Efficient Algorithms for Their Computation and Storage in Resource-Constrained Embedded Vision Systems

**Shoaib Ehsan** [1,*], **Adrian F. Clark** [1], **Naveed ur Rehman** [2] **and Klaus D. McDonald-Maier** [1]

[1] School of Computer Science and Electronic Engineering, University of Essex, Colchester CO4 3SQ, UK; E-Mails: alien@essex.ac.uk (A.F.C.); kdm@essex.ac.uk (K.D.M.-M.)

[2] Department of Electrical Engineering, COMSATS Institute of Information Technology, Islamabad 44000, Pakistan; E-Mail: naveed.rehman@comsats.edu.pk

* Author to whom correspondence should be addressed; E-Mail: sehsan@essex.ac.uk; Tel.: +44-1206-874-376; Fax: +44-1206-872-788.

**Abstract:** The *integral image*, an intermediate image representation, has found extensive use in multi-scale local feature detection algorithms, such as Speeded-Up Robust Features (SURF), allowing fast computation of rectangular features at constant speed, independent of filter size. For resource-constrained real-time embedded vision systems, computation and storage of integral image presents several design challenges due to strict timing and hardware limitations. Although calculation of the integral image only consists of simple addition operations, the total number of operations is large owing to the generally large size of image data. Recursive equations allow substantial decrease in the number of operations but require calculation in a serial fashion. This paper presents two new hardware algorithms that are based on the decomposition of these recursive equations, allowing calculation of up to four integral image values in a row-parallel way without significantly increasing the number of operations. An efficient design strategy is also proposed for a parallel integral image computation unit to reduce the size of the required internal memory (nearly 35% for common HD video). Addressing the storage problem of integral image in embedded vision systems, the paper presents two algorithms which allow substantial decrease (at least 44.44%) in the memory requirements. Finally, the paper provides a case study that highlights the utility of the proposed architectures in embedded vision systems.

## 1. Introduction

Originally proposed as the summed-area table for texture-mapping in computer graphics in the mid-1980s [1], the integral image is comparatively new in the image processing domain. The idea of using an integral image was introduced as an intermediate image representation by the Viola-Jones face detector [2]. Since then, it has been particularly useful for fast implementation of image pyramids in multi-scale computer vision algorithms such as Speeded-Up Robust Features (SURF) and Fast Approximated SIFT [3,4].

The primary reason for using an integral image is the improved execution speed for computing box filters. Employment of the integral image eliminates computationally expensive multiplications for box filter calculation, reducing it to three addition operations [2]. This allows all box filters to be computed at a constant speed, irrespective of their size; this is a major advantage for computer vision algorithms, especially feature detection techniques which utilize multi-scale analysis. Such algorithms generally require calculation of variable-size box filters to implement different scales of an image pyramid. For example, SURF requires computation of $9 \times 9$ box filters for implementation of the smallest and $195 \times 195$ for the largest scale of its image pyramid [3]; without an integral image, these larger filters would take almost 500 times longer than the smallest one to compute.

Although speed gain and reduced computational complexity are major benefits of integral image, its calculation introduces a performance overhead [5]. Image processing and computer vision algorithms are generally computation and data intensive in nature, and integral image calculation is no exception. Although it involves only additions, the total number of operations is significant due to its dependence upon the input image size. Recursive equations due to Viola and Jones [2] reduce the total number of additions but require that calculation is done in a serial fashion because of the data dependencies involved. This is not desirable for real-time embedded vision systems that have strict time limits and restricted hardware resources for processing a single frame, possibly coupled with power constraints.

Since serial calculation can provide only one integral image value per clock cycle at best, there is a strong motivation to investigate methods for efficient computation of the integral image. Indeed, there are examples in the literature where efficient computation of the integral image has been achieved on a variety of computing platforms such as multi-core processors, GPUs (Graphics Processing Units), and custom hardware [5–29]. For example, integral image calculation is accelerated by first computing the sum of all pixels in the horizontal direction and then in the vertical direction utilizing the huge computational resources of a GPU (ATI HD4850 in this particular case) in [6]. This paper also takes a step in this direction. Firstly, it performs an analysis of the recursive equations and the data dependencies involved for parallel calculation of integral image; it then proposes two hardware algorithms based on the decomposition of these recursive equations, allowing simultaneous computation of up to four integral image values in a row-parallel way without any significant increase in the number of addition operations. An efficient design strategy for a parallel integral image computation engine is then presented which reduces the internal memory requirements significantly.

Another drawback of the utilization of the integral image representation is the substantial increase in the memory requirements for its storage [30]. This is essentially due to the significantly larger word length of integral image values as compared to the original pixel values. Again, for embedded vision systems it becomes a bottleneck due to the strict constraints on hardware resources. In [30], two techniques are presented for reducing the word length of integral image: an exact method which reduces the word length by computation through the overflow without any loss of accuracy on platforms with complement-coded arithmetic; and an approximate technique which is based on rounding the input image by value truncation. The exact method is useful only when the maximum size of the box filter is considerably smaller than the input image size. Loss of accuracy is the main drawback of the approximate method. To address these issues, this paper presents two generic methods for reducing the storage requirements of the integral image significantly which can benefit both custom hardware design and software implementation on programmable processor architectures for resource constrained embedded vision systems. Finally, the paper discusses a case study to highlight the usefulness of the proposed architectures in resource-constrained embedded vision systems.

The remainder of this paper is structured as follows: an analysis of the computation of the integral image is given in Section 2. Proposed in Section 3 is a parallel computation strategy that provides two integral image values per clock cycle. Section 4 presents another parallel method that delivers four integral image values per clock cycle. Extending the approach of Section 3, a memory-efficient design strategy is proposed for a parallel integral image computation unit in Section 5. A comparative analysis of the proposed parallel methods is done in Section 6. Two methods for reducing the size of memory for storing integral image are presented in Section 7. A case study showing the utility of the proposed architectures for resource-constrained embedded vision systems is discussed in Section 8. Finally, conclusions are given in Section 9.

## 2. Analysis of Integral Image Computation

This section analyzes integral image calculation from a parallel computation perspective. The value of the integral image at any location $(x, y)$ in an image is the sum of all the pixels to the left of it and above it, including itself, as shown in Figure 1. This can be stated mathematically as in [2]:

$$ii(x,y) = \sum_{x' \leq x, y' \leq y} i(x',y') \tag{1}$$

where $ii(x, y)$ and $i(x, y)$ are the values of the integral image and the input image respectively at location $(x, y)$.

Equation (1) has potential for parallel computation, providing the input image is stored in memory and all its pixel values can be accessed. For example, the integral image of a 2 × 2 image may be computed in parallel using the following set of equations:

$$ii(1,1) = i(1,1) \tag{2}$$

$$ii(1,2) = i(1,1) + i(1,2) \tag{3}$$

$$ii(2,1) = i(1,1) + i(2,1) \qquad (4)$$

$$ii(2,2) = i(1,1) + i(1,2) + i(2,1) + i(2,2) \qquad (5)$$

Although Equation (1) can be used for small images, the number of additions involved scales as $\frac{1}{4}M^2N^2$ for an input image of size $M \times N$ pixels [5]. For example, 1,866,240,000 addition operations are required to compute the integral image for a medium resolution image of size 360 × 240 pixels. Thus, Equation (1) is not particularly suitable from a hardware perspective.

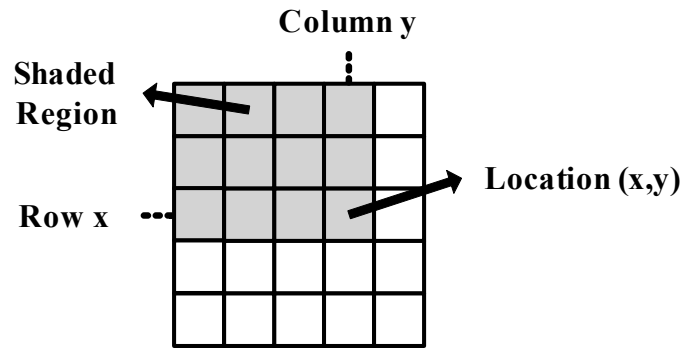

**Figure 1.** Calculation of integral image value at image location $(x, y)$. The shaded region indicates all pixels to be summed.

The total number of addition operations can be drastically reduced by utilizing the recursive equations presented in [2]:

$$S(x,y) = i(x,y) + S(x,y-1) \qquad (6)$$

$$ii(x,y) = ii(x-1,y) + S(x,y) \qquad (7)$$

where $i(x,y)$ is the input pixel value at image location $(x,y)$, $S(x,y)$ is the cumulative row sum value at image location $(x,y)$ and $ii(x,y)$ is the integral image value at image location $(x,y)$. These equations reduce the number of additions involved to $2MN$.

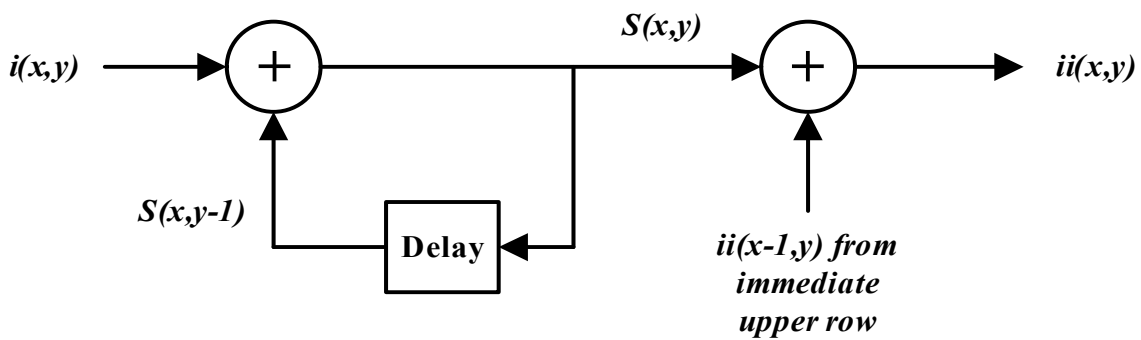

**Figure 2.** Data Flow Graph of the Viola-Jones recursive equations for a single row of the input image.

Equations (6) and (7) represent a two-stage system which operates in a serial fashion: the first stage computes the cumulative row sum at a specific image location and forwards the data to the second stage for calculation of the integral image value at that particular location. The data flow graph of this serial system is shown in Figure 2 for a single row of the input image. It can be observed from Figure 2 that individual stages are also dependent upon data from previous iterations for their operation. The first stage requires the cumulative row sum to be computed in a serial way for a single row of the input image. The second stage is more complex as it needs data from the previous row to calculate an integral image value. Hence, there is little opportunity for parallel computation in single row operations.

However, a deeper analysis of Equation (6) shows that it is possible to compute the cumulative row sum for all rows independently and hence simultaneously. This is however not true for Equation (7) due to its dependency on data from the neighboring row. Thus, the best possible system using these equations is to process individual rows in a delayed fashion. As an example, Figure 3 shows a 5 × 5 image for which integral image values are calculated by processing all rows in parallel using these equations. The shaded blocks represent the pixels for which integral image values are computed simultaneously; blocks with a cross sign indicate pixels whose integral image values have already been calculated; and white blocks show pixels for which integral image values still need to be calculated. It can easily be seen that the integral image value for the second pixel in the third row cannot be calculated until the integral image value for the second pixel in the second row is calculated. Figure 4 shows the time delay involved in computation of integral image values for different rows.

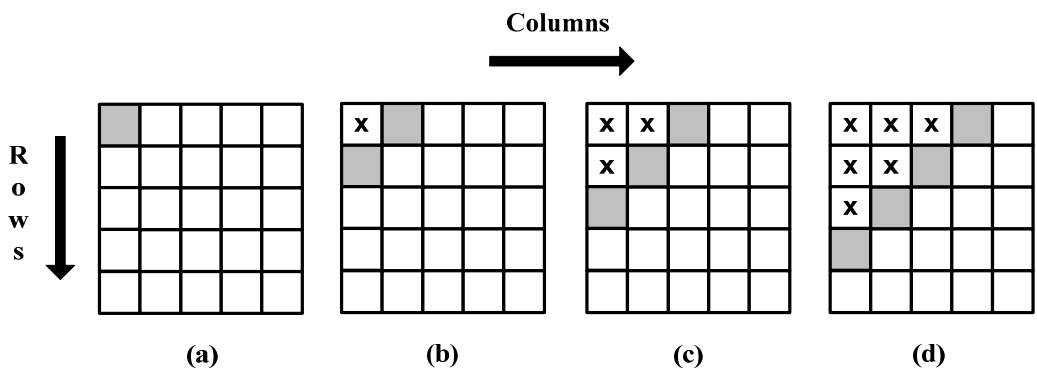

**Figure 3.** Delayed row computation using the Viola-Jones recursive equations.

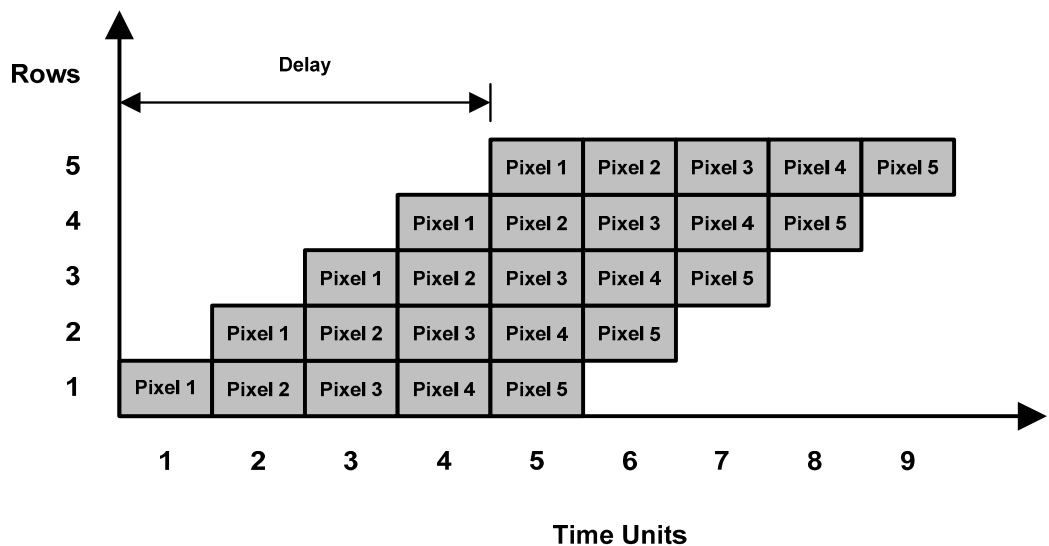

**Figure 4.** Time delay between computation of integral image values for different rows.

## 3. Parallel Computation for Two Rows

The proposed algorithm represents a two-stage, pipelined system that processes two rows of an input image in parallel, providing two integral image values per clock cycle without any delay when the pipeline is full. In particular, it allows calculation of the second pixel of the two rows in the *same* clock cycle. The whole image is divided in groups of two rows and one group is processed at a time, moving from the top to the bottom of the input image. The following set of equations is used for calculation of integral image values in a row-parallel way:

$$S(x,y) = i(x,y) + S(x,y-1) \tag{8}$$

$$S(x+1,y) = i(x+1,y) + S(x+1,y-1) \tag{9}$$

$$ii(x,y) = ii(x-1,y) + S(x,y) \tag{10}$$

$$ii(x+1,y) = ii(x-1,y) + S(x,y) + S(x+1,y) \tag{11}$$

where Equations (8) and (10) are for computation of integral image values in the first row; and Equations (9) and (11) are for the second row.

This set of equations requires $2MN + \frac{MN}{2}$ addition operations for an input image of size $M \times N$ pixels. This is not a significant increase compared to the $2MN$ additions required for the standard recursive equations, Equations (6) and (7). For all odd rows, two additions are required per pixel, as given by Equations (8) and (10). An extra addition is done for each pixel in the even rows in Equation (11) to allow simultaneous calculation of integral image values for even and odd rows without any delay. The block diagram for the proposed architecture is shown in Figure 5. A pipelined approach for this two-stage system reduces the critical data path from two adders to one. The proposed system computes the integral image in No. of Rows x No. of Columns/2 clock cycles. The execution time is governed by No. of Clock Cycles/Max Clock Frequency.

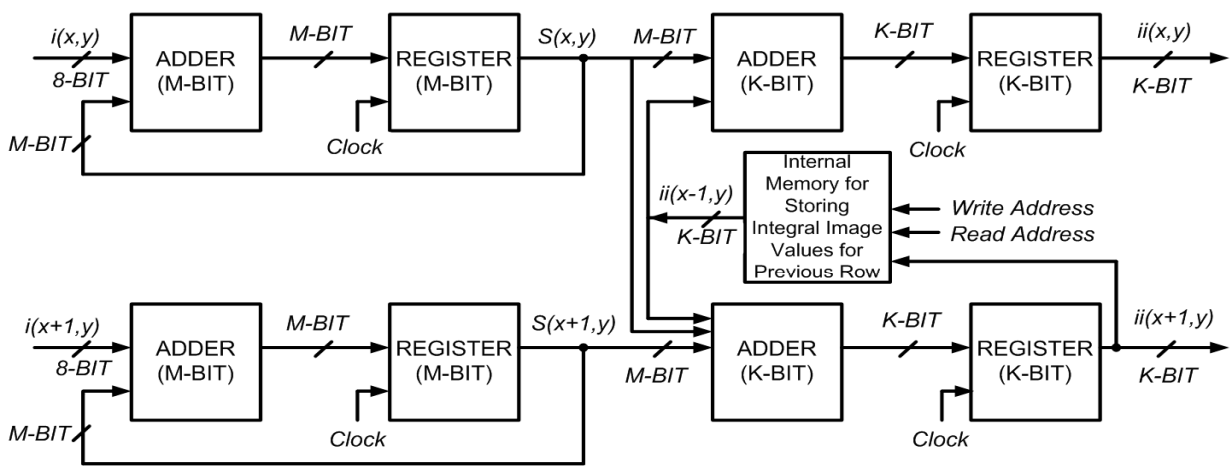

**Figure 5.** Block diagram of the proposed architecture for parallel computation of integral image for 2 rows.

## 4. Parallel Computation for Four and *n* Rows

The above algorithm for processing two rows in parallel can be extended to four rows, though at the expense of extra additions per pixel in rows 3 and 4, allowing calculation of four integral image values per clock cycle. However, this is not an attractive option as it involves more hardware. With the objective of minimizing hardware resources, another decomposition of Equations (6) and (7) is proposed in this section; it provides *four* integral image values per clock cycle in a row-parallel way with $2MN + \frac{MN}{2}$ additions for an input image of size $M \times N$ pixels.

The proposed algorithm represents a three-stage, pipelined system (as opposed to the two-stage one above) to reduce the computational resources required in hardware. It processes four rows of an input image in parallel, providing four integral image values per clock cycle. In this case, the image is divided in groups of four rows and one group is processed at a time moving from top to bottom. The following set of equations is used for calculation of integral image values in a row-parallel way:

$$S(x,y) = i(x,y) + S(x,y-1) \tag{12}$$

$$S(x+1,y) = i(x+1,y) + S(x+1,y-1) \tag{13}$$

$$S(x+2,y) = i(x+2,y) + S(x+2,y-1) \tag{14}$$

$$S(x+3,y) = i(x+3,y) + S(x+3,y-1) \tag{15}$$

$$ii(x,y) = ii(x-1,y) + S(x,y) \tag{16}$$

$$ii(x+1,y) = ii(x-1,y) + S(x,y) + S(x+1,y) \tag{17}$$

$$ii(x+2,y) = ii(x+1,y) + S(x+2,y) \tag{18}$$

$$ii(x+3,y) = ii(x+1,y) + S(x+2,y) + S(x+3,y) \tag{19}$$

where Equations (12) and (16) are for computation of integral image values in the first row; Equations (13) and (17) are for the second row; Equations (14) and (18) are for the third row; and Equations (15) and (19) are for the fourth row.

The main advantage of this system is that it requires $2MN + \frac{MN}{2}$ addition operations for an input image of size $M \times N$ pixels as is required for parallel processing of 2 rows. The block diagram for the proposed architecture is shown in Figure 6. This scheme computes the integral image in No. of Rows x No. of Columns/4 clock cycles. The execution time is governed by No. of Clock Cycles/Max Clock Frequency. It should be noted that the proposed architecture is scalable and can easily be extended to any multiple of two rows by decompositions similar to those shown above (Equation (12) to Equation (19)) for achieving more speed-up. The general model for any *n*-row architecture can be derived as:

$$S(x+j,y) = i(x+j,y) + S(x+j,y-1) \tag{20}$$

For odd rows:

$$ii(x+2k,y) = ii(x+2k-1,y) + S(x+2k,y) \tag{21}$$

For even rows:

$$ii(x + 2m + 1, y) = ii(x + 2m - 1, y) + S(x + 2m, y) + S(x + 2m + 1, y) \qquad (22)$$

where $n$ = number of rows to be computed (always a multiple of 2), $j = 0,\ldots, n-1$, $k = 0,\ldots,(\frac{n}{2} - 1)$ $m = 0,\ldots,(\frac{n}{2} - 1)$.

Naturally, this speed-up comes at the cost of more hardware resources and more power consumption. As one complete row of integral image values need to be stored for any recursion-based architecture, processing more rows in parallel at the expense of more hardware resources may be feasible for small sized images, but not for large ones. However, if there are no strict constraints on the hardware resources to be utilized and achieving more speed-up is the priority, the architecture will be viable for large image sizes as well. Since the focus in this paper is on resource-constrained embedded vision systems, we limit the discussion to 2-rows and 4-rows architectures proposed above. The detailed resource utilization and timing results for these architectures, which will be presented shortly, can be extrapolated to get estimates for any scaled *n*-rows architecture.

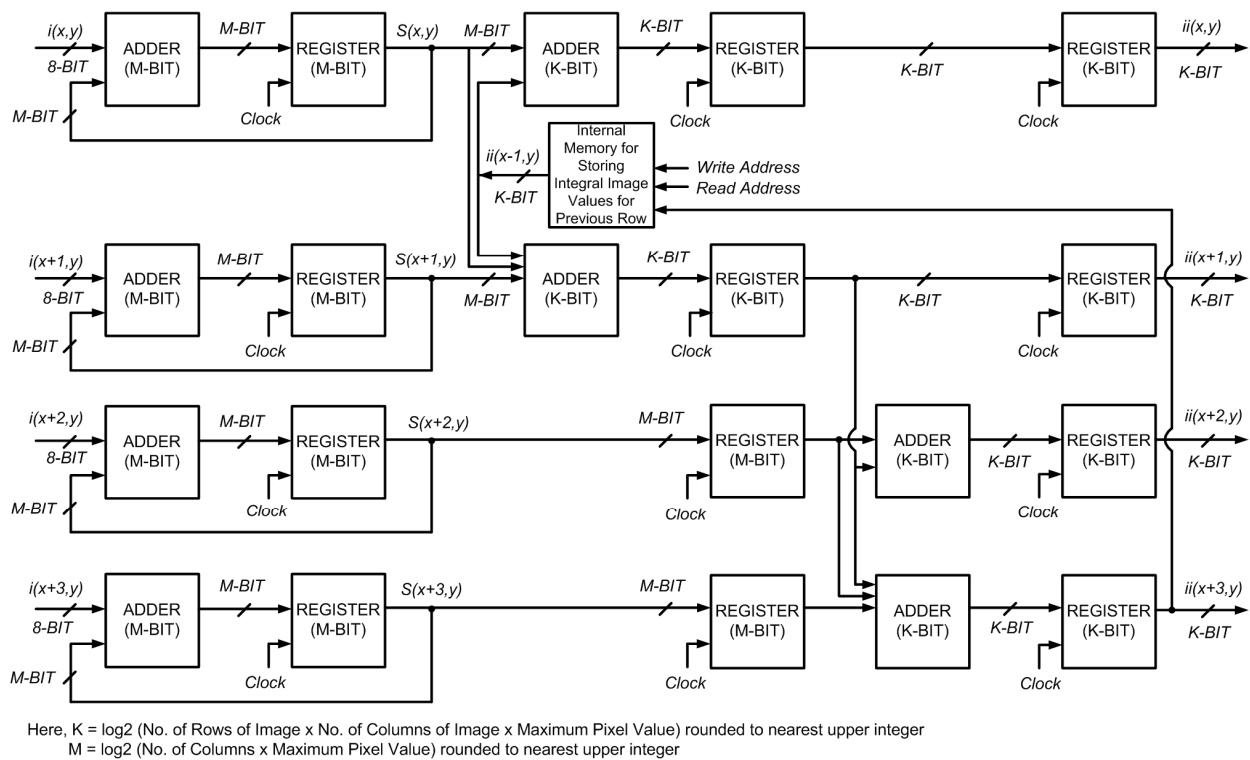

**Figure 6.** Block diagram of the proposed architecture for parallel computation of integral image for 4 rows.

Table 1 presents comparative resource utilization results for prototype implementations of the serial method (Viola-Jones recursive equations), the proposed 2-rows, and the 4-rows algorithms on a Xilinx Virtex-6 FPGA for some common image sizes with 8-bit pixels. The architectures are implemented using Verilog. All three implementations achieve a maximum frequency of about 147 MHz. Please note that the values in parenthesis in Table 1 indicate the percentage of Virtex-6 resources consumed by each specific implementation. It is evident that, without significant increase in the utilized resources, the proposed algorithms provide two and four times speed-up relative to the serial algorithm for the same image size. As all these algorithms are recursive in nature, one complete row of integral image values needs to be stored for the calculation of the very next row. This implies that these algorithms have same internal memory requirements and a major portion of the consumed resources for these techniques comes from implementing this task (see Table 1). Duplicating the number of adders between each alternative technique, therefore, has a relatively small effect on resource utilization within the same image size. As expected, the resource consumption for the three compared algorithms increases with the increasing image size and all the methods show a similar trend due to the same internal memory requirements.

**Table 1.** Comparative resource utilization results for Serial, 2-rows and 4-rows parallel prototype implementations on a Xilinx Virtex-6 XC6VLX240T FPGA for some common image sizes with 8-bit pixels. The values in parenthesis indicate the percentage of Virtex-6 resources utilized.

| Image Size | Serial | | 2-Rows Parallel | | 4-Rows Parallel | |
|---|---|---|---|---|---|---|
| | Slice Registers | LUTs (Look-Up Tables) | Slice Registers | LUTs (Look-Up Tables) | Slice Registers | LUTs (Look-Up Tables) |
| **360 × 240** | 9050 (6.00%) | 3465 (2.29%) | 9075 (6.02%) | 3606 (2.39%) | 9128 (6.05%) | 3789 (2.51%) |
| **720 × 576** | 19,488 (12.92%) | 7344 (4.87%) | 19,515 (12.94%) | 7502 (4.97%) | 19,791 (13.13%) | 7721 (5.12%) |
| **800 × 640** | 21,648 (14.36%) | 8155 (5.41%) | 21,701 (14.39%) | 8276 (5.49%) | 21,967 (14.57%) | 8506 (5.64%) |
| **1280 × 720** | 36,134 (23.97%) | 13,426 (8.90%) | 36,293 (24.07%) | 13,548 (8.98%) | 36,522 (24.23%) | 13,765 (9.13%) |
| **1920 × 1080** | 55,732 (36.97%) | 20,707 (13.73%) | 55,761 (36.99%) | 20,823 (13.81%) | 56,932 (37.77%) | 21,059 (13.97%) |
| **2048 × 1536** | 61,495 (40.80%) | 22,816 (15.13%) | 61,525 (40.82%) | 22,890 (15.18%) | 62,853 (41.70%) | 23,164 (15.36%) |
| **2048 × 2048** | 76,869 (50.98%) | 28,520 (18.91%) | 76,907 (51.01%) | 28,613 (18.97%) | 78,567 (52.11%) | 28,955 (19.20%) |

Finally, comparative timing results of integral image computation for the serial, 2-rows and 4-rows parallel prototype implementations on a Xilinx Virtex-6 FPGA are given in Table 2 for some common image sizes with 8-bit pixels. The architectures are implemented using Verilog. We have used integer format for representing the values and there is no loss of information. For the specific purpose of testing the proposed architectures and measuring the execution times, blocks of memory were instantiated in the Virtex-6 FPGA for a particular input image size and the corresponding output integral image size. The execution time was measured using Chipscope from the input of the architecture to its output. It is evident that the proposed 2-rows and 4-rows algorithms outperform serial implementation.

**Table 2.** Comparative timing results of integral image computation for Serial, 2-rows and 4-rows parallel prototype implementations on a Xilinx Virtex-6 XC6VLX240T FPGA for some common image sizes with 8-bit pixels.

| Image Size | Execution Time in Milliseconds | | |
|---|---|---|---|
| | Serial | 2-Rows Parallel | 4-Rows Parallel |
| **360 × 240** | 0.587 | 0.293 | 0.146 |
| **720 × 576** | 2.821 | 1.410 | 0.705 |
| **800 × 640** | 3.482 | 1.741 | 0.870 |
| **1280 × 720** | 6.269 | 3.134 | 1.567 |
| **1920 × 1080** | 14.106 | 7.053 | 3.526 |
| **2048 × 1536** | 21.399 | 10.699 | 5.349 |
| **2048 × 2048** | 28.532 | 14.266 | 7.133 |

## 5. A Memory-Efficient Parallel Architecture

In embedded vision systems, parallel computation of the integral image produces numerous design challenges in terms of speed, hardware resources and power consumption. Although recursive equations significantly reduce the number of operations for computing the integral image, the required internal memory becomes excessively large for an embedded integral image computation engine for increasing image sizes.

With the objective of achieving high throughput with low hardware resources, this section proposes a memory-efficient design strategy for a parallel embedded integral image computation engine. Results indicate that the design attains nearly 35% reduction in memory utilization for common HD video.

Both the recursion-based serial [2] and parallel methods (in Sections 3 and 4) require one complete row of integral image values to be stored in an internal memory so that it can be utilized for the calculation of the very next row. The width of the required internal memory is $\log_2$(number of rows × number of columns × maximum image pixel value) rounded to the upper integer whereas the depth is equal to the total number of columns in one row of the image. Figure 7 highlights the internal memory requirements for an integral image computation engine implemented in hardware for some common images sizes. It is evident that with the increasing image size, the design of the integral image computation engine becomes inefficient in terms of hardware resources due to the large internal memory. It is desirable to achieve a design which is memory-efficient and provides high throughput.

To address the internal memory problem discussed above, a resource-efficient architecture is presented that is also capable of achieving high throughput. The design strategy makes use of the fact that integral image values in adjacent columns of a single row differ by a column sum (Figure 8). This difference value is maximum in the last row as the column sum includes all pixel values from the top to the bottom of the image in a particular column. In the worst case, the difference between two adjacent columns in the last row of the image will be the product of the number of rows and the maximum value that can be attained by an image pixel (e.g., the maximum value is 255 for an 8-bit pixel).

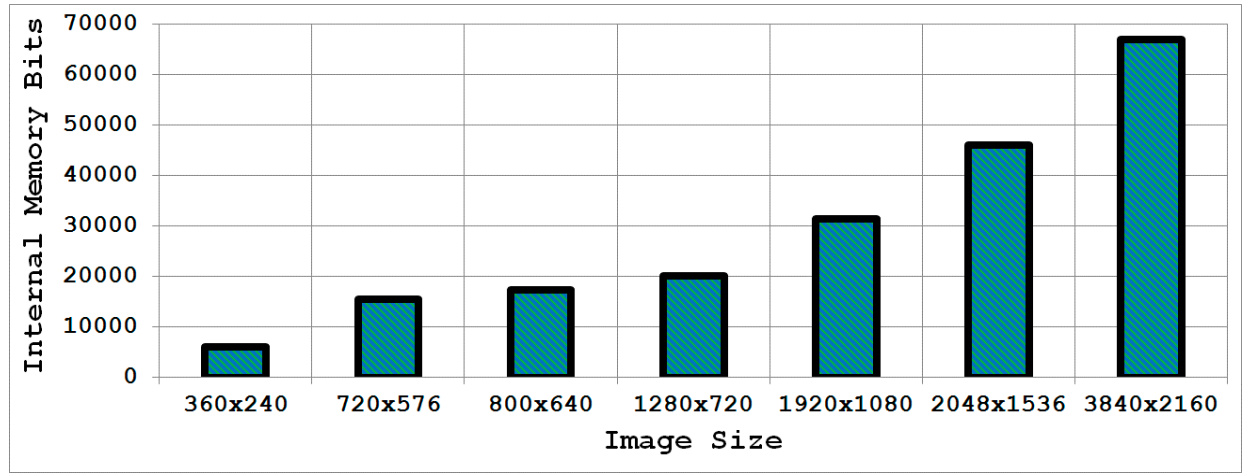

**Figure 7.** Internal memory requirements for the integral image computation engine for some common image sizes.

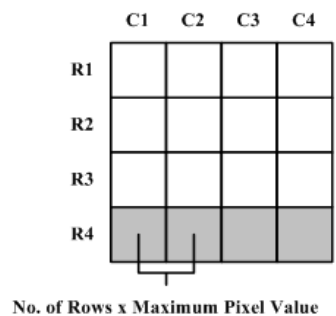

**Figure 8.** Worst case difference between adjacent integral image values in one row.

Figure 9 shows a block diagram of the proposed architecture for an embedded integral image computation engine. This pipelined architecture computes two integral image values in a single clock cycle. Unlike the parallel methods presented in Sections 3 and 4 which store a complete row of integral image values in internal memory for computing the next row, this design strategy saves only the difference values of the adjacent columns in a row for calculating the next row. Only the integral image value for the first column in that row is saved in a separate register to allow computation of the integral image values from the stored difference values. Although the depth of the internal memory remains the same as mentioned above, the proposed design approach requires the width to be $\log_2$(number of rows × maximum image pixel value) rounded to the upper integer value. Table 3 provides the timing results and the results for internal memory reduction when prototyped on an FPGA, a Virtex-6 XC6VLX240T device, for some common image sizes with 8-bit pixels. The architecture is implemented using Verilog. The maximum frequency of the design is 146.71 MHz. Please note that the values in parenthesis indicate the percentage of Virtex-6 resources consumed. It is evident from Table 3 that the architecture is capable of achieving significant memory reduction over other recursion-based methods, even for small image sizes while providing high throughput.

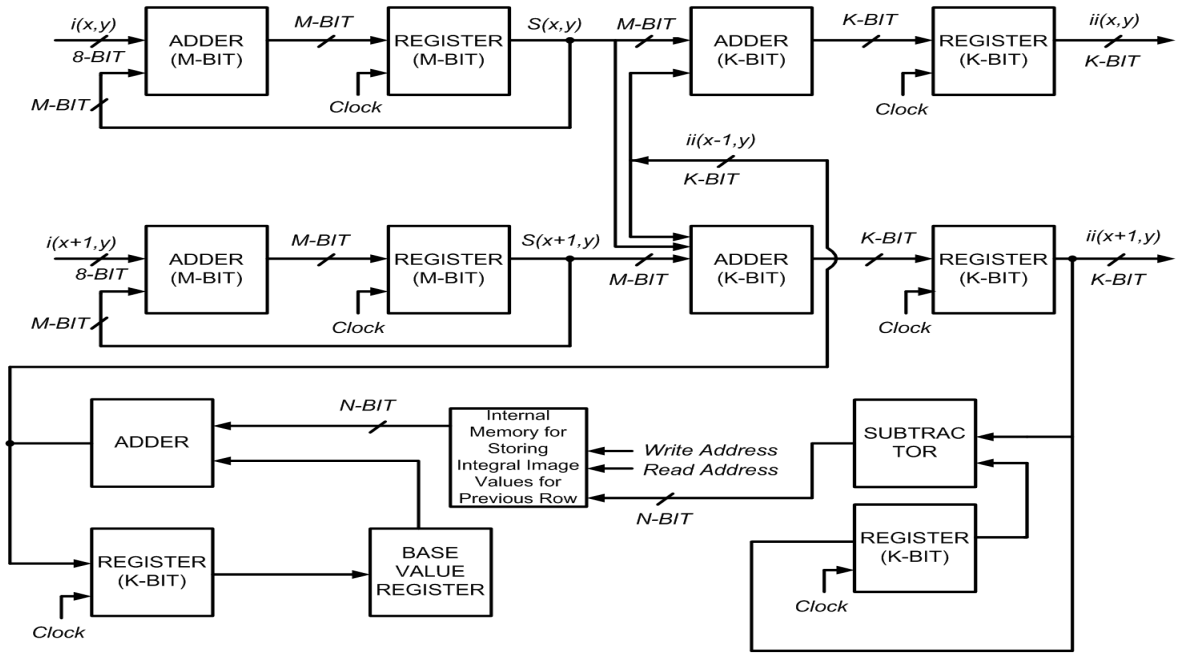

Here, K = log2 (No. of Rows of Image x No. of Columns of Image x Maximum Pixel Value) rounded to nearest upper integer
M = log2 (No. of Columns x Maximum Pixel Value) rounded to nearest upper integer
N = log2 (No. of Rows x Maximum Pixel Value) rounded to nearest upper integer

**Figure 9.** Block diagram of the proposed architecture. $i(x,y)$ and $ii(x,y)$ are the image pixel value and the integral image value at location $(x,y)$ in the image. $S(x,y)$ is the row sum at that particular location.

**Table 3.** Timing results and reduction in internal memory requirements for the proposed architecture on Virtex-6 XC6VLX240T. The values in parenthesis indicate the percentage of resources utilized for Virtex-6.

| Image Size | Memory-Efficient Design Strategy | | | Reduction in Internal Memory Bits Relative to Other Recursion Based Methods | Reduction in Resource Consumption Relative to 2-Rows Algorithm | |
|---|---|---|---|---|---|---|
| | Slice Registers | LUTs (Look-Up Tables) | Execution Time in Milliseconds | | Slice Registers | LUTs (Look-Up Tables) |
| **360 × 240** | 6307 (4.18%) | 2792 (1.85%) | 0.294 | 32% | 30.50% | 22.57% |
| **720 × 576** | 13,164 (8.73%) | 5537 (3.67%) | 1.413 | 33.3% | 32.54% | 26.19% |
| **800 × 640** | 14,602 (9.68%) | 6047 (4.01%) | 1.744 | 33.3% | 32.71% | 26.93% |
| **1280 × 720** | 24,668 (16.36%) | 9864 (6.54%) | 3.140 | 32.1% | 32.03% | 27.19% |
| **1920 × 1080** | 37,145 (24.64%) | 14,614 (9.69%) | 7.067 | 34.4% | 33.39% | 29.82% |
| **2048 × 1536** | 39,694 (26.33%) | 15,558 (10.32%) | 10.720 | 36.6% | 35.48% | 32.03% |
| **2048 × 2048** | 49618 (32.91%) | 19448 (12.89%) | 14.294 | 36.6% | 35.49% | 32.05% |

## 6. Comparative Analysis of the Proposed Algorithms with Other Implementations

Given the differences in computing technologies (multi-core processors, GPU, FPGA) and performance measures used, fair and meaningful comparisons of our proposed algorithms with other implementations is a difficult task. The contrasting application domains of the computing technologies do not help this either. Multi-core processors and GPU-based computing solutions, running at high clock rates with huge computational resources, are considered an expensive option for resource-constrained embedded systems due to their high power consumption. Although comparing the proposed algorithms with previous implementations on such computing platforms in all probability lead to biased results from an embedded system designer's viewpoint, the findings may be useful from a general computation perspective.

Performing unbiased and meaningful comparisons of our methods with equivalent previous FPGA implementations is also complex given the differences in characteristics (architecture, part and speed grade) of the FPGAs used and the image sizes considered. Moreover, integral image computation unit has usually been implemented on the FPGA as part of a larger system in the literature (e.g., [7,9–11,18–20]). Most publications only report the results for the implemented system as a whole and do not provide detailed results for resource utilization and execution time for the integral image computation unit separately (e.g., [7,9–11]). The platform-specific nature of some previous FPGA implementations also makes this assessment complicated.

Nevertheless, we here attempt useful comparisons of our proposed algorithms with other FPGA implementations reported in the literature to the extent possible. In an effort to increase the scope of this work and to make it comprehensive, our presented algorithms are also assessed rigorously by comparing against the best results achieved by some previous implementations on multi-core processors and GPUs—platforms with relatively huge hardware resources.

Before making this comparison, we would like to reiterate that the aim of this paper is to achieve fast execution with low resources—something which is critical for resource-constrained embedded vision systems. For a fair comparison, the quality of an algorithm needs to be judged both on the basis of execution time and hardware resources consumed. An algorithm that does well on both fronts is of high value.

In [6], a method is proposed for parallel computation of integral image on a GPU. The technique is implemented on ATI HD4850 GPU and the timing results are presented. The operating frequency of the graphics processor and the shaders is 625 MHz. This implementation, however, does not work well for small images. For example, it takes 27.8 ms in total to compute integral image representation for a $256 \times 256$ input image. On the other hand, the three proposed algorithms (in Sections 3–5) consume less than 0.3 ms to compute integral image of size $360 \times 240$ (please see Tables 2 and 3). Even for relatively large image sizes, our proposed algorithms outperform the GPU implementation (less than 11 ms for calculating $2048 \times 1536$ integral image for the proposed algorithms as compared to 74.1 ms for computing $2048 \times 1024$ integral image on the GPU), approximately achieving more than 7 times speed-up even with our 2-rows algorithms.

A technique is presented in [16] for fast computation of integral image using graphics hardware. The method is implemented on three different GPUs: Radeon 9800 XT (412 MHz graphics processor), Radeon X800XT PE (500 MHz graphics processor) and GeForce 6800 Ultra (400 MHz graphics processor). The best results are achieved by Radeon X800XT PE. This GPU computes a 1024 × 1024 integral image in 36.2 ms. It is evident from Tables 2 and 3 that the three proposed algorithms require significantly smaller execution time for calculating much larger integral images (more than 9 times speed-up for the 2-rows algorithms; more than 18 times speed-up for the 4-rows algorithm).

In [25], integral image computation is implemented on an NVIDIA GeForce GTX 295 graphics card. The clock frequency of the graphics processor is 576 MHz, whereas the shaders operate at 1242 MHz. It is shown that computation of a 2048 × 2048 integral image takes about 5.5 ms for this GPU implementation. Extrapolating the results given in Tables 2 and 3 for a 2048 × 2048 integral image indicates that the fastest of the three proposed algorithms (4-rows algorithm) would require 7.133 ms to do this computation. Although the GPU implementation [25] achieves better execution time, the performance of the proposed algorithms is still commendable given the huge hardware resources gap between the computing technologies used.

Similarly, timing results for integral image calculation on an NVIDIA GeForce 9600GT card are given in [29]. The operating frequency of the graphics processor is 650 MHz, whereas the shaders are clocked at a rate of 1625 MHz. This implementation takes 9.562 ms to compute a 2048 × 2048 integral image. Our proposed 4-rows algorithm requires 7.133 ms while the remaining two presented techniques consume less than 14.3 ms for the same computation. Again, the performance of our algorithms is laudable considering that these techniques are targeted for resource-constrained embedded vision systems.

Three different multi-core CPUs are utilized in [13] to implement four different integral image computation methods: T8100 running at 2.1 GHz, P8600 at 2.4 GHz, and E5405 at 2.0 GHz. It is reported that E5405 achieves the best execution time (about 4 ms for 2048 × 1536 image size). Our proposed 4-rows algorithm achieves comparable execution time of 5.349 ms with significantly less hardware resources. The other two presented methods also perform well (less than 11 ms).

A parallel implementation of integral image is reported in [27] on a Tile64 MIMD-based multi-core system running at 750 MHz. For an image size of 1920 × 1080, this implementation takes 0.11 second to compute the integral image. On the other hand, our proposed algorithms perform much better in terms of execution speed for the same image size (please see Tables 2 and 3), achieving at least 15 times speed-up for the three proposed algorithms. The presented algorithms also outperform two different implementations of integral image reported in [5], that utilized recursion and double buffering techniques, on a 600 MHz media processor TMS320DM6437 which consume 11.2 ms and 1.8 ms respectively for computing integral image of size 720 × 480.

An integral image computation unit is implemented as part of a hardware architecture for face detection using feature cascade classifiers on a Xilinx Virtex-II Pro XC2VP30 FPGA in [7]. The maximum clock frequency for the architecture is 126.8 MHz. This Xilinx-specific FPGA implementation consumes 20901 LUTs for the whole architecture. No separate resource utilization and timing results are reported for the integral image computation unit in [7], which makes the comparison difficult. An integral image computation architecture based on systolic arrays is presented in [26]. The architecture is scalable and operates at 144.07 MHz on a Virtex-6 FPGA. The integral image is calculated in a delayed row fashion (as shown in Figure 3). For achieving nearly four-times speed-up with respect to the serial algorithm, this architecture consumes 41363 LUTs for an image of size 640 × 480. An equivalent FPGA implementation of our 4-rows algorithm consumes at least five-times less LUTs as compared to [26] for a much larger 800 × 640 image (see Table 1).

In [18], an integral image unit is implemented on a Virtex-II Pro FPGA as part of a network-on-chip architecture for face detection. This unit consumes 8590 gates for an image of size 320 × 240, and the maximum operating frequency is 163.074 MHz. No information is available about execution time for the integral image unit. It is therefore difficult to make a comparison with our algorithms in this case. For an image size of 352 × 288, an integral image unit is implemented in [19] on a Virtex-II FPGA as part of a people detection system. The architecture operates at 100 MHz and the execution time is 30 ms. No separate resource utilization details are available for the integral image unit. Our algorithms outperform this implementation in terms of execution time (Tables 2 and 3).

In this section, we have shown relevant comparisons of our algorithms with previous implementations found in the literature in a meaningful way. It is evident that our proposed algorithms not only outperform previous FPGA implementations but also fare exceptionally well when compared against multi-core and GPU implementations that utilize relatively huge computational resources.

## 7. Efficient Storage of Integral Image

As opposed to its computation, storage of the integral image has received less attention until recently. The only work of significance in this domain is presented in [30]. Memory requirements for an integral image are substantially larger than for the input image. For resource-constrained embedded vision systems, storage of the integral image presents several design challenges. In this section, two viable techniques for reducing the memory requirements of an integral image are proposed for different application scenarios. Both hardware and software solutions can benefit from the presented techniques. Results for some common image sizes are presented which show that the methods guarantee a minimum of 44.44% reduction in memory for all image sizes and application scenarios, and may achieve reduction of more than 50% in specific situations for embedded vision systems.

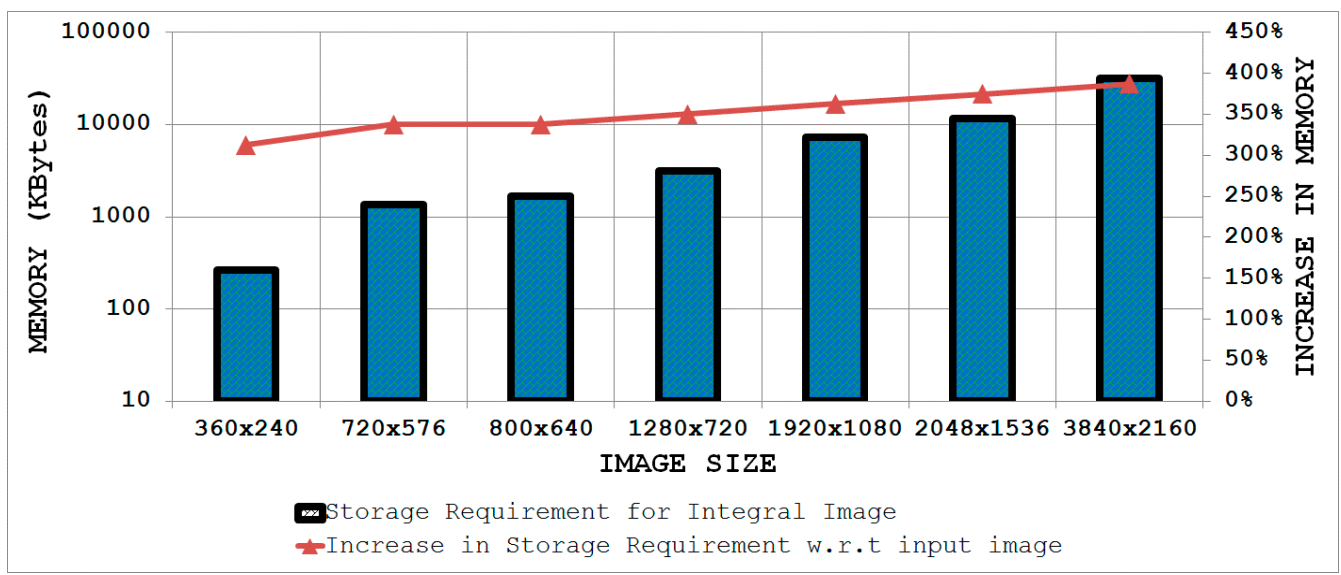

**Figure 10.** Storage requirements of the integral image for some common image sizes and percentage increase in memory relative to the input image (considering 8-bit pixels).

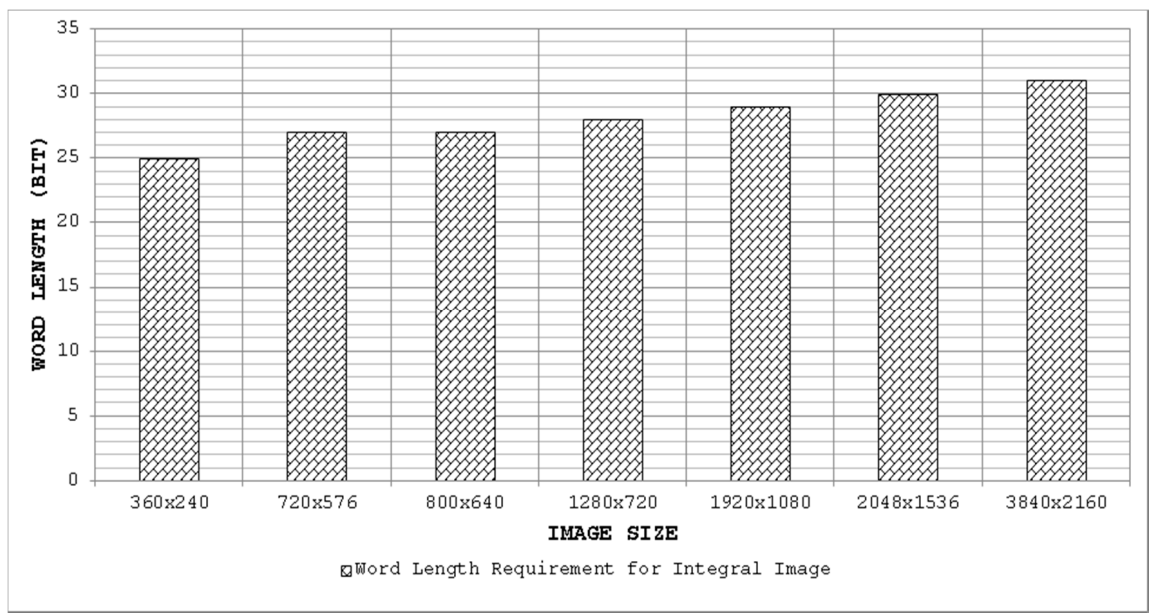

**Figure 11.** Word length requirements for integral image for some common image sizes considering 8-bit input pixels.

The bars in Figure 10 show the storage requirements of an integral image for some common image sizes (read values from the left ordinate axis), while the line indicates the percentage increase in memory with respect to the input image considering 8-bit pixels (read values from the right ordinate axis). It is evident from Figure 10 that the storage requirements wide). Figure 11 depicts the word lengths required for an integral image considering 8-bit input pixel values for some common image sizes. It can be seen clearly that with increasing image size, the required word length for the integral image also increases.

*7.1. Limitations of Existing Methods*

Although the exact and approximate methods presented in [30] manage to reduce the word length of an integral image, they do have some limitations:

(a) These methods are applicable only in situations where the size of the box filter is a priori known.
(b) The exact method achieves negligible reduction in memory if the maximum size of the box filter is almost equal to the input image size.
(c) The approximate method involves loss of accuracy due to rounding pixel values. For example, there is significant increase in false detection rate for the Viola-Jones face detector when the approximate method is used in [30].
(d) Although the exact method does not incur any loss of accuracy, it fails to achieve significant reduction in word length.
(e) These techniques are one-dimensional in the sense that they only concentrate only on reducing the word length of the integral image, which in turn affects the width of the storage memory but not its depth.

To overcome the above-mentioned shortcomings, two methods are presented for storing the integral image efficiently in embedded vision systems without any loss of accuracy. The first of these is appropriate for any application that involves an integral image without prior knowledge of the box filter size and in situations where the size of the box filter is nearly the same as that of the input image. The second method is suited to applications where the size of box filter is a priori known (e.g., SURF [3]).

*7.2. Proposed Method 1*

This is a general technique that guarantees 44.44% reduction in memory resources for storing an integral image and can be utilized for any application involving integral images. The method is lossless and is suitable for scenarios where the box filter size is either unknown or is not much smaller than the image size. The technique is especially attractive for embedded systems, as the same system can be utilized for different applications without any modifications to hardware or software.

**Figure 12.** A sample 3 × 3 integral image block for the proposed method. The shaded region shows the integral image values that need to be stored.

Unlike the methods in [30], the proposed technique attempts to reduce the depth of the memory required to store an integral image. For this particular method, the width of the memory is assumed to be $log_2$(length of the image × width of the image x maximum pixel value) rounded to the upper integer value. The first step is to make the length and width of the integral image both into multiples of 3. For example, if the integral image dimensions are 360 × 240 then the length and the width values are already multiples of 3 and nothing needs to be done. Otherwise, the last rows and/or columns of the integral image are discarded to achieve this objective. In the worst case, the last two rows and the last two columns need to be eliminated. The whole integral image is then divided into blocks of 3 × 3 integral image values. Figure 12 depicts a single such block. The shaded integral images values in Figure 12 are the ones that are selected by the proposed method to store in the memory; the remaining four values on the corners are discarded. Despite not storing these four corner integral image values, the 3 × 3 integral image block can be perfectly reconstructed from the stored integral image values by utilizing the fact that:

$$a = b + d - e + input\ pixel\ value\ at\ e \tag{23}$$

$$c = b + f - e - input\ pixel\ value\ at\ f \tag{24}$$

$$g = d + h - e - input\ pixel\ value\ at\ h \tag{25}$$

$$i = h + f - e + input\ pixel\ value\ at\ i \tag{26}$$

Figure 13 shows all 3 × 3 blocks for a sample integral image of dimensions 9 × 9 (with values shaded as to whether they need to be stored or discarded). Out of the 81 integral image values in Figure 13, only 45 values need to be stored in memory, meaning that this method achieves a 44.44% reduction in the storage requirements for the integral image. Moreover, this reduction is independent of the input image size and the box filter size.

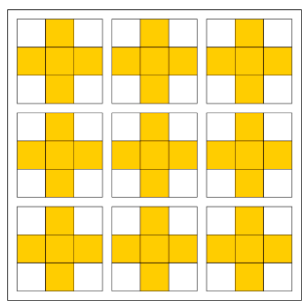

**Figure 13.** A sample integral image of dimensions 9 × 9. The shaded regions indicate the integral image values that need to be stored in the memory.

As a box type filter can be computed quickly using three addition and subtraction operations when the integral image values on the four corners of that filter are known [2] (see Figure 14), the proposed method does not require any extra computation if the required four values are those which are stored in memory. In the worst case, all four integral image values needed for computing the box filter will not be available from memory. In that particular case, Equation (23) to Equation (26) can be utilized for computing the integral image values which were discarded earlier; they can then be used for calculating the required box filter. Although there is a speed-memory tradeoff involved, the method is still an efficient way of computing box type filters as it eliminates computation intensive multiplications.

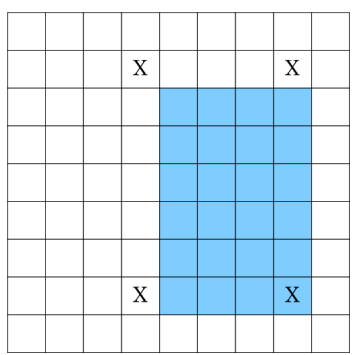

**Figure 14.** Box filter calculation using the integral image; the shaded area indicates the filter to be computed whereas 'X' shows the integral image values required for computation of this box filter.

*7.3. Proposed Method 2*

In an effort to reduce the size of the memory required for storing the integral image further, a technique is presented here which decreases both the width and the depth of memory. It combines the exact method presented in [30] with the technique proposed in Section 7.2.

This hybrid method is suitable for scenarios where the maximum size of the box filter to be computed is considerably smaller than the input image size. Again, the method does not incur any loss of accuracy.

The worst-case integral image value that determines the binary word length required to represent integral image is dependent upon the width, height and number of bits per pixel of the input image. This can be stated as [30]:

$$ii_{max} = (2^{L_i} - 1) \times W \times H \tag{27}$$

where $i$ is the input image, $ii$ is the integral image, $ii_{max}$ is the worst case integral image value, $W$ is the width of the input image, $H$ is the height of the input image and $L_i$ is the numberof bits per pixel for the input image. According to [30], the number of bits $L_{ii}$ required for representing the worst case integral image value thus needs to satisfy:

$$(2^{L_{ii}} - 1) \geq (2^{L_i} - 1) \times W \times H \tag{28}$$

The total memory in bytes required to store the integral image can be calculated as follows:

$$Memory = \frac{(W \times H) \times L_{ii}}{8} \tag{29}$$

According to the exact method in [30], for platforms with complement-coded arithmetic, if the maximum height and the width of the box filter to be calculated are known, then the word length for the integral image needs to satisfy:

$$(2^{L_{ii}} - 1) \geq (2^{L_i} - 1) \times W_{max} \times H_{max} \tag{30}$$

where $W_{max}$ is the maximum width of box filter and $H_{max}$ is the maximum height of a box filter. Equation (30) can be explained on the basis that if a chain of linear operations is performed on integers and there are some intermediate overflowing results then it is possible to get the correct final result if this result can be represented by the used data word length [30]. The proposed method first makes both the length and width of the input image multiples of 3. Equation (30) is then used to find the required word length for storing the integral image. As a final step, the depth of the memory is reduced by employing the method proposed in Section 7.2.

A variant of this method can also prove useful. Observing Equation (30) closely reveals that the supposition of having all pixel values in the input image set to their maximum value (255 in the case of 8-bit pixels) for evaluating the worst case integral image value does not seem very practical for feature detection techniques like SURF which is used for blob detection. *i.e.*, to detect dark areas/regions in the input image surrounded by light ones or *vice versa*. Assuming that all the pixels are set to their maximum value in the input image implies that there is absolutely no variation in the pixel values. Since most feature detection techniques try to detect features in those areas of the image where there are large changes in pixel values, this assumption simply means that there are no features to be detected in the input image.

This variant of the above technique further extends the exact method of [30] by supposing that there is variation in pixel values of the input image. Equation (30) is thus modified as:

$$(2^{L_{ii}} - 1) \geq [(2^{L_i} - 1) \times (W_{max} \times H_{max}) \times 0.96$$
$$+(2^{L_i-1} - 1) \times (W_{max} \times H_{max}) \times 0.04] \quad (31)$$

It is assumed here that 96% of all pixels in a box filter to be evaluated have maximum values, while the other 4% of pixels have half the maximum value. This is a suitable approximation as most images generally have more variation in pixel values than given by Equation (31). The final step is to reduce the depth of the required memory by employing the technique presented in Section 7.2.

Given the results presented in Figure 11 for the word length with the increasing image size, it would be interesting to compare our proposed methods, not only with [30], but also with some floating-point representation (that utilizes less than 32 bits) for storing the integral image. Although integer representations may have a limited range as compared to floating-point representations for a given word length, both can represent equal number of distinct values. For integer representations, the spacing between the numbers is equal. However, for floating-point representations, the distances between numbers are denser when the number is small, and sparser when the number is large. Thus, the absolute representation error increases with larger numbers for the floating point representations. For example, the IEEE 754-2008 half-precision floating point format (16-bit), for which the maximum value that can be represented is 65,504, has the following precision limitations:

(a) Integers between 0 and 2048 can be exactly represented.
(b) Integers between 2049 and 4096 round to a multiple of 2.
(c) Integers between 4097 and 8192 round to a multiple of 4.
(d) Integers between 8193 and 16,384 round to a multiple of 8.
(e) Integers between 16,385 and 32,768 round to a multiple of 16.
(f) Integers between 32,769 and 65,536 round to a multiple of 32.

For the particular case of integral image, we are only dealing with the whole numbers. In an effort to reduce the required memory, representing integral image values using a small word length floating-point representation may increase the range as compared to integer representations, but will lead to large representation and computation errors with increasing values as shown above.

Nevertheless, to highlight the usefulness of our methods, we have selected a customized 17-bit floating-point representation for comparison. This representation is utilized on the basis that the 32-bit single-precision IEEE 754 format (having one sign bit, 8 exponent bits and 23 significand bits) can represent a maximum value of $\approx 3.4 \times 10^{38}$, much more than what is required even for fairly large-sized integral images. The single-precision format can easily represent the worst case integral image value for an image of size 3840 × 2160 (which is $\approx 2.115 \times 10^9$) considered here. On the other hand, the 16-bit half-precision floating-point format (with 1 sign bit, 5 exponent bits and 10 significand bits) can only represent a maximum value of 65,504 (suitable for the worst case integral image value of a small image patch of 16 × 16). We have, therefore, added one more bit to the exponent of the half-precision floating point format to create a customized 17-bit floating-point format that can represent a maximum value of $(2 - 2^{-10}) \times 2^{31} \approx 4.29 \times 10^9$ (making it suitable for a 3840 × 2160 image).

Figure 15 shows comparative results for the two variants of the proposed method, the customized 17-bit floating-point representation, and the original exact method [30] for the specific case of the SURF detector with increasing image sizes by taking $W_{max} = 65$ and $H_{max} = 129$. Note that the largest box filter to be computed for SURF is 195 × 195 but it can be broken down into three box type filters of 65 × 129 (or 129 × 65) [3]. The bars in Figure 15 represent the memory required for storing the integral image (read values from the left ordinate axis) whereas the line graphs show the percentage reduction in memory (read values from the right ordinate axis) relative to the actual requirement (see Figure 10). It is evident that the best performance in terms of memory reduction is achieved by utilizing Equation (31) in combination with the depth reduction method from Section 7.2 (Variant 2 in Figure 15). It can be seen that the two variants of the proposed method out-perform the original exact method and the 17-bit floating-point representation comprehensively and allow more than 50% reduction in memory, even for small sized images.

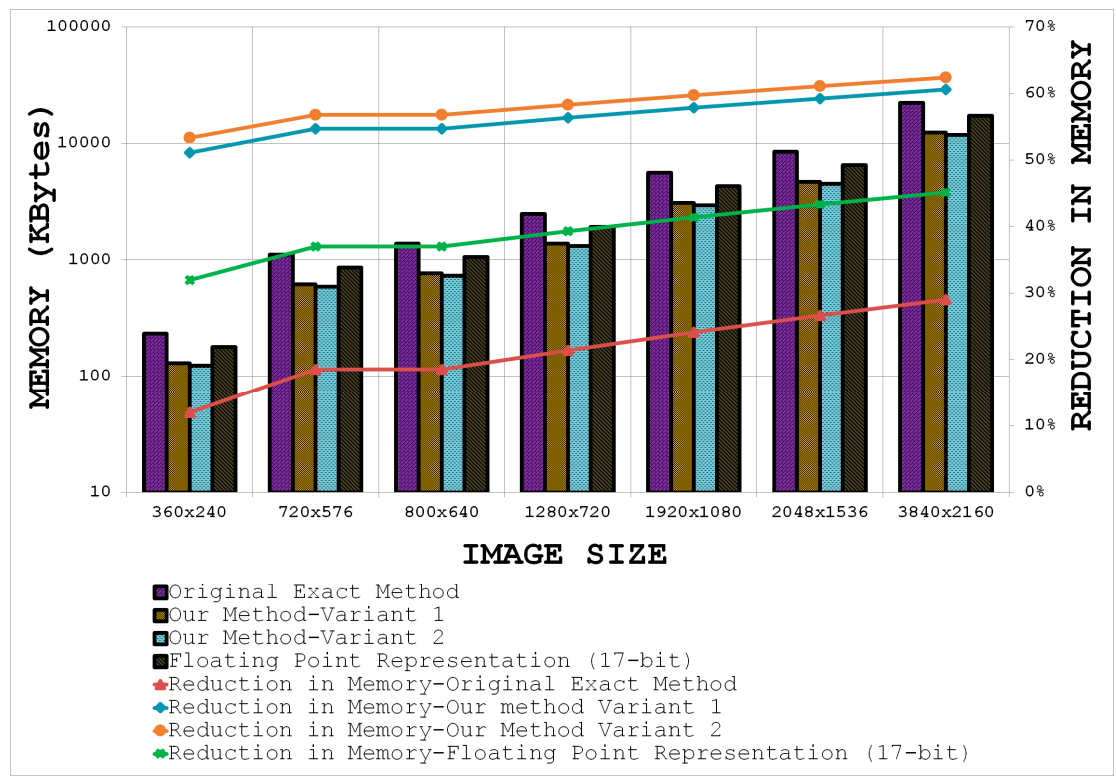

**Figure 15.** Comparative results for the original exact method [30] and the two variants of the proposed technique.

## 8. A Case Study

This section gives an example to illustrate the impact of the proposed architectures on some real application scenarios. Adaptive document binarization is a key primary step in many document analysis and OCR processes [31]. To show the usefulness of the proposed integral image computation architectures, an FPGA implementation of the fast adaptive binarization algorithm for greyscale documents (presented in [32]) is exposed as a case study here. This fast adaptive binarization algorithm yields the same quality of binarization as the Sauvola method [31], but executes in time close to that of global thresholding methods (such as Otsu's method [33]), independent of the window size. This algorithm combines the statistical constraints of Sauvola's method with integral images. For more details, please see [32].

Figure 16 shows a block diagram of our FPGA implementation for the fast adaptive binarization algorithm [32] utilizing the proposed four-rows integral image computation architecture. The system is implemented on a Xilinx Virtex-6 FPGA. The integral image computation is done using our four-rows architecture in hardware, while the remainder of the fast adaptive binarization algorithm [32] is implemented in software on a Xilinx MicroBlaze processor. We have used a local window size of 15 × 15 for the implementation of the algorithm [32]. The resource utilization results for the implemented system and its execution time for different image sizes are reported and compared in Table 4 with the following: (1) a software only implementation of [32] on a Xilinx MicroBlaze processor and (2) the hardware implementation of the serial integral image algorithm combined with a software implementation of the remainder of the fast adaptive binarization algorithm [32]. Since all the three considered implementations execute the same code for fast adaptive binarization algorithm [32] on a MicroBlaze processor, it allows a fair comparison between the three implemented systems to analyze the effects of different integral image computation schemes on the overall system performance.

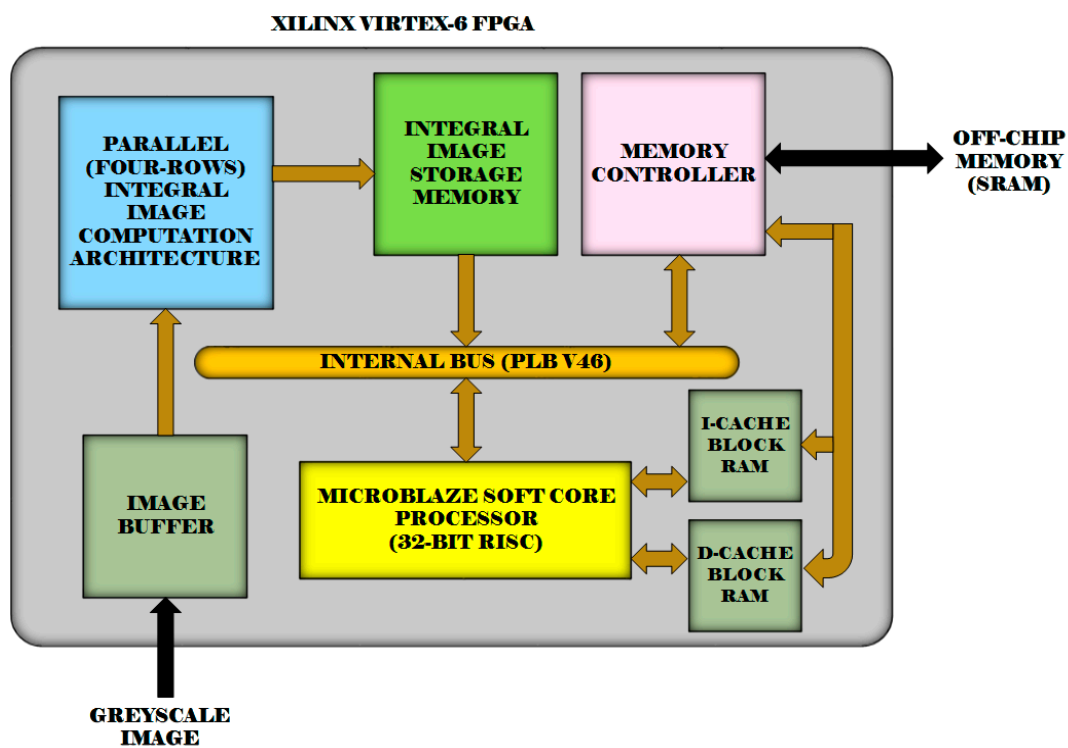

**Figure 16.** A block diagram of the implemented fast adaptive binarization system [32] on a Xilinx Virtex-6 FPGA.

Clearly, with a small increase in the utilized FPGA resources relative to the other two implementations, the implementation utilizing our proposed architectures achieves approximately 3.6 times and 5.3 times speed up with respect to the serial hardware implementation and the software-only implementation on a MicroBlaze processor. It is therefore evident from the results presented in Table 4 that our proposed architectures serve as useful building blocks for larger embedded vision systems for real-world vision sensing application scenarios and offer substantially enhanced performance.

**Table 4.** Comparative FPGA resource utilization and timing results for the three prototype implementations of the fast adaptive binarization algorithm [32] on a Xilinx Virtex-6 XC6VLX240T FPGA for some common image sizes with 8-bit pixels. The values in parenthesis indicate the percentage of Virtex-6 resources utilized.

| Image Size | Software-Only Implementation of [32] Executing on MicroBlaze Processor Implemented on Virtex-6 | | Serial Integral Image Computation in Hardware + Software Implementation of the Rest of the Algorithm [32] on MicroBlaze Processor Implemented on Virtex-6 | | 4-Rows Parallel Integral Image Computation in Hardware + Software Implementation of the Rest of the Algorithm [32] on MicroBlaze Processor Implemented on Virtex-6 | |
|---|---|---|---|---|---|---|
| | LUTs (Look-Up Tables) | Execution Time (s) | LUTs (Look-Up Tables) | Execution Time (s) | LUTs (Look-Up Tables) | Execution Time (s) |
| **360 × 240** | 8503 (5.61%) | 0.43 | 11,968 (7.9%) | 0.29 | 12,292 (8.12%) | 0.08 |
| **720 × 576** | 40,814 (26.97%) | 2.07 | 48,158 (31.82%) | 1.34 | 48,535 (32.07%) | 0.38 |
| **800 × 640** | 50,387 (33.3%) | 2.55 | 58,542 (38.69%) | 1.65 | 58,893 (38.92%) | 0.46 |
| **1280 × 720** | 71,253 (47.09%) | 4.59 | 79,479 (52.52%) | 2.97 | 79,981 (52.85%) | 0.82 |
| **1920 × 1080** | 92,271 (60.98%) | 10.32 | 100,893 (66.67%) | 6.68 | 101,345 (66.97%) | 1.84 |
| **2048 × 1536** | 115,926 (76.61%) | 15.65 | 123,348 (81.51%) | 10.08 | 123,869 (81.86%) | 2.75 |
| **2048 × 2048** | 129,113 (85.32%) | 20.81 | 138,657 (91.63%) | 13.40 | 139,057 (91.90%) | 3.69 |

To further analyze and highlight the significance of the integral image computation block for the fast adaptive binarization algorithm [32], Figure 17 shows the comparative timing analysis in logarithmic scale for the three implemented systems on a Xilinx Virtex-6 FPGA for some common image sizes. It gives the timing breakdown of the three implementations and clearly highlights how an optimized integral image computation block positively affects the whole computation and makes a substantial performance difference. It is evident from Figure 17 that slow integral image computation for the software-only implementation has a substantial negative effect on the overall system performance in terms of computation time. On the other hand, the system with hardware implementation of the serial integral image computation algorithm performs relatively better and also reduces the overall system computation time. Finally, it can be clearly seen from Figure 17 that the system employing our 4-rows parallel hardware for integral image computation reduces the overall system computation time significantly due to the optimized integral image computation block. To highlight this further, Figure 18 shows the comparative system throughput results in logarithmic scale (in frames/second) for the three implementations for some common image sizes. Again, the system utilizing our proposed 4-rows parallel integral image computation hardware achieves substantially enhanced throughput as compared to the other two implementations, thus showing the utility of our architectures as building blocks for larger embedded vision systems.

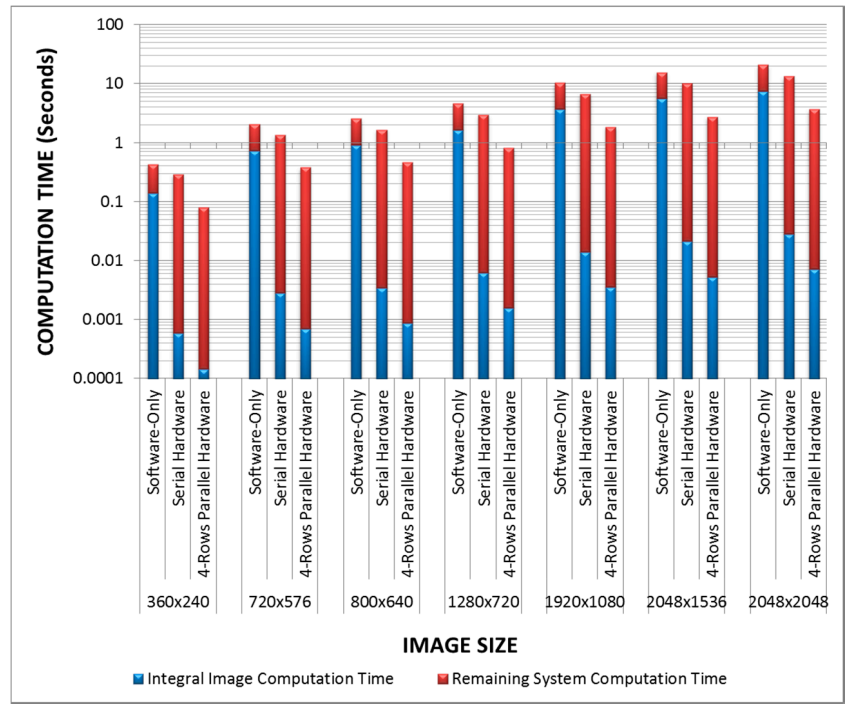

**Figure 17.** Comparative timing analysis (in logarithmic scale) of the three prototype implementations of the fast adaptive binarization algorithm [32] on a Xilinx Virtex-6 FPGA for some common image sizes.

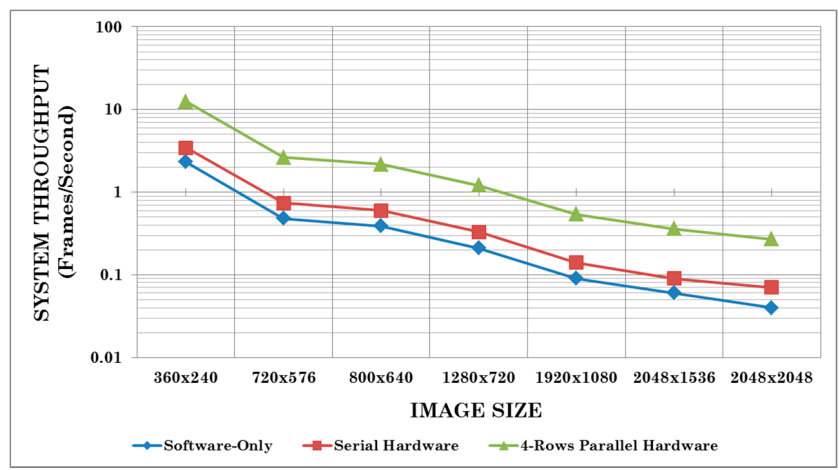

**Figure 18.** Comparative system throughput results in logarithmic scale (in frames/second) for the three prototype implementations of the fast adaptive binarization algorithm [32] on a Xilinx Virtex-6 FPGA for some common image sizes.

## 9. Conclusions

This paper has addressed computation and storage issues related to integral images. It has analyzed integral image calculation from a parallel computation perspective. With the objective of reducing computational resources, two hardware algorithms based on the decomposition of the Viola-Jones recursive equations are proposed in this paper. These are capable of providing up to four integral image values per clock cycle without any significant increase in the number of addition operations. An efficient design strategy for a parallel embedded integral image computation engine that is capable of achieving nearly 35% reduction in internal memory for common HD video (1920 × 1080) was also proposed. The paper has presented two methods for the reduction of memory for storing an integral image. These techniques guarantee at least 44.44% reduction in memory and may allow more than 50% reduction when the maximum size of a box filter to be computed is considerably smaller than the input image size. Finally, the paper provides a case study that highlights the usefulness of the proposed architectures.

The paper has primarily focused on integral image computation and storage in resource-constrained embedded vision systems because these are what are generally found in mobile robots etc. At the moment, such systems do not typically employ high-dynamic range (HDR) images, though it is possible that they may do so in a few years' time. Of course, the principles that led to the algorithms expounded in the paper can clearly be applied to HDR images too and is a promising future direction. Also, the proposed algorithms are essentially parallel and so have the potential of being implemented on platforms other than FPGA, providing they have appropriate parallel computing resources (such as a GPU). This idea can be explored further in future.

## Acknowledgments

The authors would like to thank the anonymous reviewers for their positive comments, helpful suggestions and constructive feedback. This work was supported by the UK EPSRC under Grant EP/K004638/1 (RoBoSAS).

## Author Contributions

S. Ehsan, A.F. Clark and K. D. McDonald-Maier conceived and designed the algorithms and the architectures; S. Ehsan and N. Rehman performed the experiments and analyzed the data.

## Conflicts of Interest

The authors declare no conflict of interest.

## References

1. Crow, F. Summed-area tables for texture mapping. *ACM SIGGRAPH Comput. Graph.* **1984**, *18*, 207–212.
2. Viola, P.; Jones, M. Rapid Object Detection using a Boosted Cascade of Simple Features. In Proceedings of the IEEE Computer Society Conference on Computer Vision and Pattern Recognition, Kauai, HI, USA, 8–14 December 2001; pp. 511–518.
3. Bay, H.; Ess, A.; Tuytelaars, T.; Gool, L. Speeded-Up Robust Features (SURF). *Comput. Vis. Image Underst.* **2008**, *110*, 346–359.
4. Grabner, M.; Grabner, H.; Bischof, H. Fast Approximated SIFT. In Proceedings of the Asian Conference on Computer Vision, Hyderabad, India, 13–16 January 2006; pp. 918–927.
5. Kisacanin, B. Integral Image Optimizations for Embedded Vision Applications. In Proceedings of the IEEE Southwest Symposium on Image Analysis and Interpretation, Santa Fe, NM, USA, 24–26 March 2008; pp. 181–184.

6. Messom, C.; Barczak, A. Stream Processing of Integral Images for Real-Time Object Detection. In Proceedings of the Ninth International Conference on Parallel and Distributed Computing, Applications and Technologies, Otago, New Zealand, 1–4 December 2008; pp. 405–412.
7. Lai, H.C.; Savvides, M.; Chen, T. Proposed FPGA Hardware Architecture for High Frame Rate (>100 fps) Face Detection using Feature Cascade Classifiers. In Proceedings of the First IEEE International Conference on Biometrics: Theory, Applications and Systems, Crystal City, VA, USA, 27–29 September 2007.
8. Zhang, N. A Novel Parallel Prefix Sum Algorithm and its Implementation on Multi-Core Platforms. In Proceedings of the Second International Conference on Computer Engineering and Technology, 2010; pp. 66–70.
9. Hiromoto, M.; Nakahara, K.; Sugano, H.; Nakamura, Y.; Miyamoto, R. A Specialized Processor Suitable for AdaBoost-based Detection with Haar-like Features. In Proceedings of the IEEE Conference on Computer Vision and Pattern Recognition, Minneapolis, MN, USA, 17–22 June 2007.
10. Gao, C.; Lu, S. Novel FPGA based Haar Classifier Face Detection Algorithm Acceleration. In Proceedings of the International Conference on Field Programmable Logic and Applications, Heidelberg, Germany, 8–10 September 2008; pp. 373–378.
11. Wei, Y.; Bing, X.; Chareonsak, C. FPGA Implementation of AdaBoost Algorithm for Detection of Face Biometrics. In Proceedings of the IEEE International Workshop on Biomedical Circuits and Systems, Singapore, 1–3 December 2004.
12. Theocharides, T.; Vijaykrishnan, N.; Irwin, M. A Parallel Architecture for Hardware Face Detection. In Proceedings of the IEEE Computer Society Annual Symposium on Emerging VLSI Technologies and Architectures, 2006.
13. Zhang, N. Working towards Efficient Parallel Computing of Integral Images on Multi-Core Processors. In Proceedings of the Second International Conference on Computer Engineering and Technology, Chengdu, China, 16–18 April 2010; pp. 30–34.
14. Hensley, J.; Scheuermann, T.; Singh, M.; Lastra, A. Interactive Summed-Area Table Generation for Glossy Environmental Reflections. In Proceedings of ACM SIGGRAPH, 2005. Available online: http://citeseerx.ist.psu.edu/viewdoc/download?doi=10.1.1.90.8836&rep=rep1&type=pdf (accessed on 10 July 2015).
15. Cho, J.; Mirzaei, S.; Oberg, J.; Kastner, R. FPGA-Based Face Detection System using Haar Classifiers. In Proceedings of the ACM/SIGDA International Symposium on Field Programmable Gate Arrays, Monterey, CA, USA, 2009; pp. 103–112.
16. Hensley, J.; Scheuermann, T.; Coombe, G.; Singh, M.; Lastra, A. Fast Summed-Area Table Generation and its Applications. *Comput. Graph. Forum* **2005**, *24*, 547–555.
17. Yang, M.; Wu, Y.; Crenshaw, J.; Augustine, B.; Mareachen, R. Face Detection for Automatic Exposure Control in Handheld Camera. In Proceedings of the Fourth IEEE International Conference on Computer Vision Systems, New York, NY, USA, 4–7 January 2006.
18. Lai, H.C.; Marculescu, R.; Savvides, M.; Chen, T. Communication-Aware Face Detection using NOC Architecture. In Proceedings of the Sixth International Conference on Computer Vision Systems, Greece, 12–15 May 2008; pp. 181–189.
19. Nair, V.; Laprise, P.O.; Clark, J. An FPGA-based People Detection System. *EURASIP J. Appl. Signal Process.* **2005**, *2005*, 1047–1061.

20. Ngo, H.T.; Tompkins, R.C.; Foytik, J.; Asari, V.K. An Area Efficient Modular Architecture for Real-Time Detection of Multiple Faces in Video Stream. In Proceedings of the Sixth International Conference on Information, Communications and Signal Processing, Singapore, 10–13 December 2007.
21. Sengupta, S.; Lefohn, A.E.; Owens, J. A Work-Efficient Step-Efficient Prefix Sum Algorithm. In Proceedings of the Workshop on Edge Computing Using New Commodity Architectures, Chapel Hill, North Carolina, 23–24 May 2006.
22. Blelloch, G.E. Prefix Sums and their Applications. In *Synthesis of Parallel Algorithms*; Reif, J.H., Ed.; Morgan Kaufmann Publishers Inc.: San Francisco, CA, USA, 1990; pp. 35–60.
23. Horn, D. Stream Reduction Operations for GPGPU Applications. In *GPU Gems*; Pharr, M., Ed.; Addison Wesley: Boston, MA, USA, 2005; pp. 573–589.
24. Sato, Y.; Sugimura, T.; Noda, H.; Okuno, Y.; Arimoto, K.; Nagasaki, T. Integral-Image based Implementation of U-SURF Algorithm for Embedded Super Parallel Processor. In Proceedings of the International Symposium on Intelligent Signal Processing and Communication Systems, Kanazawa, Japan, 7–9 January 2009; pp. 485–488.
25. Bilgic, B.; Horn, B.K. P.; Masaki, I. Efficient Integral Image Computation on the GPU. In Proceedings of the IEEE Intelligent Vehicles Symposium, San Diego, CA, USA, 21–24 June 2010; pp. 528–533.
26. Juan, A. Field-Programmable Gate Array Implementation of a Scalable Integral Image Architecture Based on Systolic Arrays. Master of Science Thesis, Utah State University, Logan, UT, USA, 2011.
27. Wu, Y.T.; Cho, C.Y.; Tseng, S.Y.; Liu, C.N.; King, C.T. Parallel Integral Image Generation Algorithm on Multi-Core System. In Proceedings of the 9th IEEE International Symposium on Parallel and Distributed Processing with Applications, Busan, Korea, 26–28 May 2011; pp. 31–35.
28. Terriberry, T.B.; French, L.M.; Helmsen, J. GPU Accelerating Speeded-Up Robust Features. In Proceedings of 3DPVT, Atlanta, GA, USA, 18–20 June 2008; pp. 355–362.
29. Huang, W.; Wu, L.; Zhang, Y. GPU-Based Computation of the Integral Image. In Proceddings of 2011 International Conference on Virtual Reality and Visualization, Beijing, China, 4–5 November 2011; pp. 243–246.
30. Belt, H. Word Length Reduction for the Integral Image. In Proceedings of the 15th IEEE International Conference on Image Processing, San Diego, CA, USA, 12–15 October 2008; pp. 805–808.
31. Sauvola, J.; Pietikainen, M. Adaptive Document Image Binarization. *Pattern Recognit.* **2000**, *33*, 225–236.
32. Shafait, F.; Keysers, D.; Breuel, T. Efficient Implementation of Local Adaptive Thresholding Techniques Using Integral Images. In Proceedings of the 15th Document Recognition and Retrieval Conference (DRR-2008), Part of the IS&T/SPIE International Symposium on Electronic Imaging, San Jose, CA, USA, 26–31 January 2008.
33. Otsu, N. A Threshold Selection Method from Gray-Level Histograms. *IEEE Trans. Syst. Man Cybern.* **1979**, *9*, 62–66.